# A Novel Approach to Develop a New Hybrid Technique for Trademark Image Retrieval

Saurabh Agarwal[1] and Punit Kumar Johari[2]

[1] [2] Department of CSE/IT,
Madhav Institute of Technology and Science, Gwalior

## Abstract

*Trademark Image Retrieval is playing a vital role as a part of CBIR System. Trademark is of great significance because it carries the status value of any company. To retrieve such a fake or copied trademark we design a retrieval system which is based on hybrid techniques. It contains a mixture of two different feature vector which combined together to give a suitable retrieval system. In the proposed system we extract the corner feature which is applied on an edge pixel image. This feature is used to extract the relevant image and to more purify the result we apply other feature which is the invariant moment feature. From the experimental results we conclude that the system is 85 percent efficient.*

## 1. Introduction

The rapid increase in the field of computer technology and digital system will help the user to store multimedia information, digital images and other digital data in an effective and processed manner. With the use of digital storage the amount of data has increased and it is a difficult task to search and get the desired outcome from this huge volume of data. As it is very tricky task for a user to search for desired needs, so to overcome this problem a demand for the retrieval system which understands the user demands and search for the required results. But to design such a system which is close enough to the human perception is a typical task.

As by the demand towards this innovative retrieval system, various researchers were attracted towards it and to work for this active research area. There were various factors to judge the overall performance of the system like the quality of the output, the time required for performing any individual query and the major factor is the difference between human perception and retrieval system must be as low as it can.

The early retrieval system uses the textual annotation. This system works on the principal of employing individual keywords to each image, and for searching the desired result the textual queries are applied in the system. This system is known as Text Based Image Retrieval System. It works well under a low amount of data, but as the data increases it become a very tough task to annotate a text or keyword for each individual. So this system is not suitable for today's scenario.

To overcome the problem of Text Based Image Retrieval system, a new system is introduced which work on content features of the image. In 1992 a new term is introduced in the field of retrieval system by Kato [1] which uses the content features, this system is well known as Content Based Image Retrieval (CBIR) system. Kato emphasized on the use of color and shape as the content feature of the image for performing the retrieval process. Later a new feature namely texture feature is also added in the field of CBIR systems.

CBIR approach is based on Query by example approach in this a query image is passed through the retrieval system and the similar images from the image database are selected which are close to query image features. The CBIR uses three main content features:

## 1.1    Shape

Shape [2] as a feature doesn't refer to the shape of any object; it refers to the properties related to shape like foreground, background, region, contour etc. From these properties the contour detection and the region detection is more popular.

## 1.2    Color

Color [3] is the easiest and closest feature with the human perception. As in this the machine also categorizes the feature and intensity value as the human does so we can say it is very close to human perception. In this the machine categorizes the images into standard color formats like RGB, CMY, HSV etc. In Color format the feature were stored according to the intensity values of the standard color which lies between 0 and 255. These intensity values were used to find the relevant images.

## 1.3    Texture

Texture [4] refers to as the repeated pattern in an image. In this two major works were performed first is to find the region which has texture pattern and then to find the properties of that visual patterns. The properties which define the texture patterns are the property of the surface having homogeneous patterns. The main features of texture are contrast, roughness, directionality, energy, entropy etc, these features were also known as the tamura [5] features.

In CBIR system the shape feature were found more flexible and accurate as compared to the other two. Because shape features are much like human observation so it is very popular between the researchers.

Trademark Image Retrieval (TIR) [6, 7] system is of great importance now-a-days. As the trademark holds the prestigious value of the company so it is very important to avoid the copying of the similar image for another company. TIR is a branch of shape base CBIR system so it is easy to build up a TIR system using the feature of shape. Trademark can be broadly classified into four different types [8]. First category is word in mark it only contains the words and character. The other one is device mark which contains specific shapes and graphical designs. The next is composite mark which is a combination of the previous two i.e. it contains both words as well as the graphical designs. The last one is complex marks it is the extension on composite mark as it consist of three dimensional graphical designs. The classification can be better understandable with the help of Figure 1.

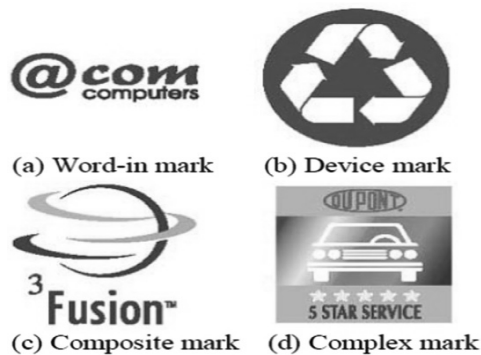

Figure1. Types of trademark (Kim & Kim, 1998)

## 2. EXISTING RETRIEVAL SYSTEM

In CBIR system the work mainly perform on the shape contents. For extracting the shape feature different shape descriptors techniques were used. The techniques were broadly classified into two main categories, one is the contour based shape descriptors and another one is region based shape descriptors.

Contour refers to the boundary pixel of any object in an image. Using the feature of boundary pixel many other contour descriptors were developed like histogram of centroid distance [9], tangential direction of contour points [10] and many more. To perform all this we need an edge boundary and it is a typical task to find a smooth and connected edge of a noisy image. To find such an edge holding both the properties is very tough but there were so many algorithms developed which nearly find a satisfactory result i.e. Canny, Sobel, Prewitt, Roberts, Prompt edge detection system.

Region refers to as the area internally covered by the edge pixel including the edge line. There are so many region based shape descriptors, some of them which are frequently used by the researchers are hu's invariant moment , Zernike moment , Wavelet transform, Fourier Descriptor, SIFT etc. Out of these we mainly emphasizes on hu's invariant moment because it has the property to handle TRS (Translation, Rotation and Scaling) structures.

There were so many previous work performed on Trademark retrieval system. Trademark retrieval is categorized in three different types of system [11] in which the active researchers are working. First from these category is TRADEMARK system which is introduced in 1990 by Kato et al. This system works on those shape descriptors which are derived from graphical shape vectors. The other system is named as STAR system and it is introduced by Wu et al. in 1996. It works on the base of CBIR system having some having some extended features of different region based shape descriptors. The last one is ARTISAN system it is introduced by Eakins et al. in the year 1996. It works on the principle of Gestalt. The Gestalt theory [12] states that the human visual perception is more conditional to the properties of image. This theory is introduced in 19th century by the team of psychologists, according to them there remain a challenge of finding accurate features.

## 3. OVERVIEW OF THE PROPOSED WORK

The proposed system will work on the principal of CBIR system. It consists of two phases i.e. offline and online phase. This combination of offline and online process can be more understandable with the help of the figure shown in Figure 2.

In the first phase which is the offline phase contains a dataset of different formats of images which is passed through a pre- processing unit which apply the function to make image mare desirable to human inputs. This step includes the changing of color formats or managing the size of image or any other pre processing functions. After applying all these function we need to find the feature of the image which may be anything depend on the applied algorithm. These features were now stored in a database for further processing on demand by the user. This whole process is performed in an offline mode i.e. the time complexity of the system doesn't depend on this process.

The other phase which is online phase is the main part or better to say the heart of the system. It is much more similar to the offline system because it has some same functions as that in the first phase. In this the user passes the query image which goes through the pre- processing and feature extraction phase these phases are exactly same as that of the offline phase. But now the main part of the unit which is the similarity measurement functions. In this the difference between the inputs of both the phases are compared to find the close common image. These extracted images were the Relevant Images which is the output of the retrieval system.

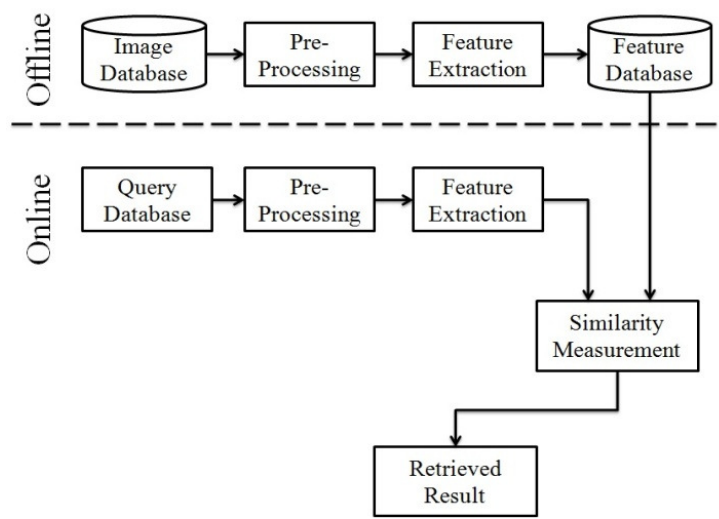

Figure 2. Image Retrieval system

## 4. FEATURE EXTRACTION METHODOLOGY

Feature extraction is a very important part of the retrieval system. The features are those points which define whole or part of an image which can be use to find the relevant images from database images. To extract the feature we use the shape descriptors, as we discussed earlier that shape descriptors are of two types out of this our main focus is on region based shape descriptors. In region based descriptor we find that corner count feature perform well, but by performing some experiments we conclude that it is not an easy task to find the corner points of a noisy or a roughly scanned image. As we are performing our experimental setup on trademark images and most of them were scanned images of different old company's logo. To extract the fine and appropriate corners in the image we must take help of Contour based shape features. After performing some of the experiment we find that prompt based edge detection finds a fine and appropriate edge of any noisy image.

## 4.1    Edge Detection

We are using Prompt based edge detection [13]. For finding appropriate edge pixel we evaluate every pixel of image one after the other. To take decision that the pixel is edge pixel or not the system performs some calculation like calculating the difference between the intensity values with its neighbouring pixels. This process helps the system for taking decisions. The elaborated process of the Prompt based edge detection is shown in the Algorithm 1.

**Algorithm 1.** Prompt Based Edge Detection

1.  Select the input image I.
2.  Find the image size in row an column form
        [R, C]= size (I);
3.  For each pixel in the image, Repeat step 4 to 6
4.  Calculate the absolute difference between all the 8 neighboring pixels.
5.  Find the number of difference that exceeds the local threshold (T).
            If,        difference > T
            Then,    k (difference count) =k+1
6.  If,        3<k<6
      Then,    the above pixel is an edge pixel.
7.  Connect all the calculated edge pixels in a single image to obtain the desired result.

## 4.2    Corner Point Detection

It refers to those points which have high changing differences with respect to their neighbouring pixels. To evaluate the corner pixels most researchers use the eigenvectors. These eigenvectors are used to build a corner matrix. It is first introduced by Harris and Stephens [14], they use the sum squared difference between the eigenvectors to find the corner pixels. For having the clearer picture of corner point detection the algorithm is shown in Algorithm 2.

**Algorithm 2.** Corner Count in an image

1.  Select the input image I.
2.  Generate the corner metric matrix of the image I.
            CM=cornermetric (I);
3.  Find the corner peaks in the CM matrix.
            (x, y) = Corner Index.
4.  Plot all the corner coordinates in the image.
5.  Calculate the total no. of corner in the image.

**4.3     Invariant moment**

In 1962, hu presented seven invariant moments [15] which are calculated for two dimensional graphical images. It is introduced for the process of pattern recognition of visual images. It is more likely to be popular between the researchers because of its flexible nature to deal with translated, rotated and scaled images.

The seven moments introduced by hu is shown below:

$\emptyset 1 = \eta 20 + \eta 02$
$\emptyset 2 = (\eta 20 - \eta 02)2 + 4\eta 211$
$\emptyset 3 = (\eta 30 - 3 \eta 12)2 + 3(\eta 21 - \eta 03)2$
$\emptyset 4 = (\eta 30 - \eta 12)2 + (\eta 21 + \eta 03)2$
$\emptyset 5 = (\eta 30 - 3 \eta 12) (\eta 30 + \eta 12) [(\eta 30 + \eta 12)2 - 3(\eta 21 + \eta 03)2] + (3 \eta 21 - \eta 03) (\eta 21 + \eta 03) [3 (\eta 30 + \eta 12)2 - (\eta 21 + \eta 03)2]$
$\emptyset 6 = (\eta 20 - \eta 02) [(\eta 30 + \eta 12)2 - (\eta 21 + \eta 03)2] + 4 \eta 11 (\eta 30 + \eta 12) (\eta 21 + \eta 03)$
$\emptyset 7 = (3 \eta 21 - \eta 03) (\eta 30 + \eta 12) [(\eta 30 + \eta 12)2 - 3(\eta 21 + \eta 03)2] + (\eta 30 - 3 \eta 12) (\eta 21 + \eta 03) [3 (\eta 30 + \eta 12)2 - (\eta 21 + \eta 03)2]$

According to the experiment performed we have a decision to make that the central moments were more reliable to handle translation invariance structures and the first two or three were more flexible with the rotational structures. To more understand the working principle of moment invariant the algorithm is shown in Algorithm 3.

**Algorithm 3.** Invariant Moment

1.  Select the input image I.
2.  Transform the image into two dimensional, real valued and numeric forms.
3.  Calculate the value of raw moment's $m_{pq}$.

$$m_{pq} = \sum_{x} \sum_{y} x^p y^q f(x, y)$$

4.  Calculate the central moment $\mu_{pq}$.

$$\mu_{pq} = \sum_{x} \sum_{y} (x - \bar{x})^p (y - \bar{y})^q f(x, y)$$

Where, $\bar{x} = \frac{m_{10}}{m_{00}}$

$\bar{y} = \frac{m_{01}}{m_{00}}$

5.  Find the normalized central moment $\eta_{pq}$.

$$\eta_{pq} = \frac{\mu_{pq}}{\mu_{00}^{\gamma}}$$

Where, $\gamma = \frac{p + q}{2} + 1$

6.  Evaluate the values of all seven hu's moments using output from step 5.

# 5. FEATURE MATCHING

The term feature matching refers to similarity measurement between the query image and the images stored in the databases. It is a very important part of the retrieval process, a good choice of matching strategy can help a system to give better and faster results and vice versa.

Normally the feature matching finds the difference between the two feature points and these differences were passed through a threshold system which filters out the unwanted result. The most commonly used feature matching system by the researchers is Euclidean distance [16] method. The Equation for calculation using Euclidean distance is shown in equation (a). In this method the squared sum of all the feature points are passed through a square root function which gives the distance calculation between the two images.

$$D(x, y) = \sqrt[2]{\sum (x_i - y_i)^2}$$  ………… (a)

### 5.1 Threshold function

It is a tough task to eliminate the relevant images from the non relevant ones. For this threshold function is used to filter out the final result. In our proposed algorithm we have main focus on the threshold system. As by the experiment performed on the retrieval system we conclude that for matching the corner points feature we have to manipulate the threshold values according to the query image. The relation between corner count and threshold value is that they are directly proportional to each other. This relation can be better understood by equation (b).

$$\text{Threshold} \propto \text{Corner Count}$$  ………… (b)

For this we design a threshold system which suits our query, for this the minimum threshold and maximum threshold are set on run time. To better understand the system please refers the Algorithm 4.

**Algorithm 4.** Threshold function
1. Find the number of corner in an image.
   Count = Cornercount (I);
2. Initialize the value of range difference coefficient R and threshold difference coefficient T.
3. For Count in range from init_R (initially 0) to final_R, repeat step 4 to 5.
4. Set, Threshold = T;
   and, T = T * Multiplying Coefficient;
5. Set, init_R = init_R + R;
   and, final_R = final_R +R;
6. Calculate the minimum and maximum threshold.
   Min_T = Count – Threshold;
   Max_T = Count + Threshold;

## 6. PROPOSED ALGORITHM

In the proposed algorithm we first apply the prompt edge detection method on the images to extract the boundary pixels. Now over target is to find that which of these boundary pixels belongs to the set of corner pixel, for this we apply the corner point detection so that we get the corner count for each individual image. Using these corner count values we find the similar images with that of the query images. To get more purify result we pass the output to the Rotational Invariant filter. The working algorithm of the whole process is shown in Algorithm 5.

**Algorithm 5. Proposed Algorithm**

1. Select the input image I.

      I = Query image

2. Convert the image in gray scale intensity values.

      Input_image = rgb2gray (I);

3. Find the Edge pixel image using Prompt edge detection.

      Edge_image=Prompt_edge (Input_image);

4. Find the corner points of the Edge pixel image.

      Corner_count = corner_point (Edge_image);

5. Apply the similarity measurement algorithm

      Difference_value=|Corner_count–Corner_count_database |

6. Find the rotational moment value of QI images (i.e. query image and the images obtained from step 5)

      Phi = invmoments (QI);

7. Display the images filtered through Step 6.

The flow Chart of the proposed algorithm is shown in Figure 3.

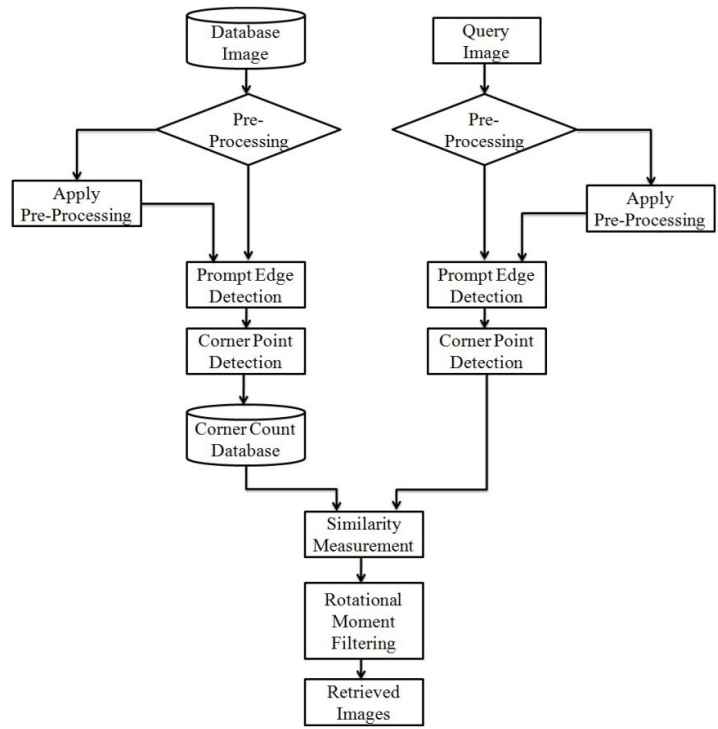

Figure 3. Flow chart of the proposed retrieval system

# 7. PERFORMANCE EVALUATION

This section displays the result obtained in different stage under the testing phase of the system. To develop such a system which satisfies the human needs is the final destination of the retrieval process. For the judgment of result with the desired goal we use the Precision and Recall graph. Precision/recall graph is the most commonly used decision making system for Trademark image retrieval system. There exists a standard formula for calculating precision and recall [17] values of a system. The formula used in the proposed experiment for evaluating the value of precision is shown in equation (c) and for recall is displayed in equation (d).

$$\text{Precision} = \frac{N_r}{T_r} \qquad\qquad \text{............ (c)}$$

$$\text{Recall} = \frac{N_r}{T_s} \qquad\qquad \text{............ (d)}$$

Where,

$N_r$ = Number of similar images in the retrieved result.
$T_r$ = Total number of images in the retrieved result.
$T_s$ = Total number of similar images in the database.

For testing phase we use a Trademark Dataset [18] of approx 108 images which has images to test rotational challenges in the system. The trademark database consist of 18 different classes of images each of which contains rotated images in six different angles i.e. 0, $\pi/3$, $2\pi/3$, $3\pi/3$, $4\pi/3$ and $5\pi/3$.

Table1. Retrieved images with their precision/recall value

As shown in Table 1. we conclude that the precision value obtained by the proposed system is 100 percent and the recall value is also nearly equal to 100 percent except for some of the query images. The overall performance of the system is found to be approx. 85 percent which is satisfactory result.

We have design a hybrid system in which first feature is used to find the relevant image and the other feature is used for filtering the result. If we individually use the two features then the result is found to be 60 percent which is not good in comparison 85 percent. The progress graph of the three systems i.e. using corner only, using invariant moment only and hybrid of the two is shown in figure 4.

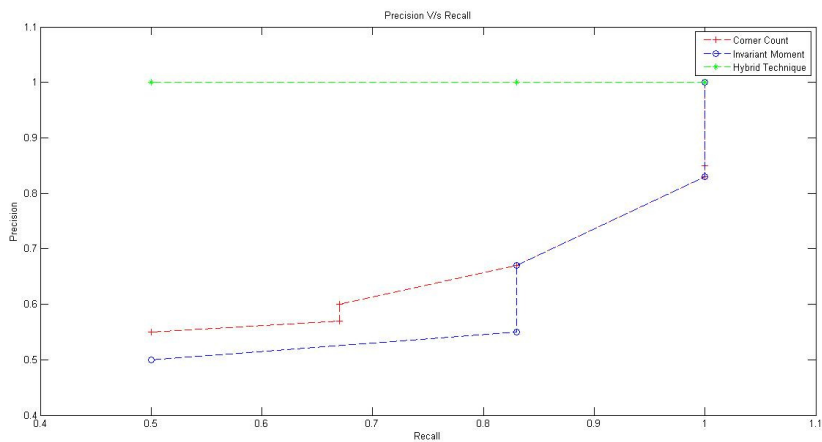

Figure 4. Precision/Recall Graph of Comparative methods

# 8. CONCLUSION AND FUTURE WORK

In the proposed work, we have design an efficient trademark retrieval system which works on the principle of CBIR system. In this system we apply a two phase feature matching strategy. One for global shape features and another is for local shape features. Different forms of transformational challenges are applied to test the efficiency of the system. We have applied the final testing on a rotational transformed image dataset. For evaluating the performance of the system we have use the precision and recall values. It is shown in the P-R graph that the system performance was satisfactory.

In future we are trying to more generalize the system so that it may handle all the other transformations. It is also very important to design such a trademark system which can handle all the types of trademarks. We can also use better clustering and efficient filtering approaches. We can use the feedback mechanism to attain such a system which is very close to human perception.

# ACKNOWLEDGEMENTS

The authors would like to thank the anonymous reviewers for their constructive comments.

# AUTHORS

**Saurabh Agarwal,** male, is currently an M.Tech. Student at Madhav Institute of Technology and Science, Gwalior, India. He got his bachelor degree from Laxmi Narayan Institute of Technology, Gwalior, India in 2012. His research interest includes digital image processing and pattern recognition. 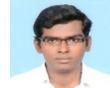

**Punit Kumar Johari**, male, is an Assistant Professor at Madhav Institute of Technology and Science, Gwalior, India. He has a working experience of 10 years in different colleges. His research interest includes Digital image processing, Pattern recognition and Data mining.